\documentclass[runningheads,a4paper]{llncs}

\usepackage{amssymb}
\usepackage{graphicx}

\usepackage{url}
\urldef{\mailsa}\path|julian.faraone@sydney.edu.au|    
\newcommand{\keywords}[1]{\par\addvspace\baselineskip
\noindent\keywordname\enspace\ignorespaces#1}

\usepackage{times}
\usepackage{graphicx} 
\usepackage{subfigure} 
\usepackage{amscd}
\usepackage{amsmath}
\usepackage{amssymb}

\usepackage{amsthm}
\usepackage{mathtools}

\usepackage{epsfig}
\usepackage{epstopdf}
\usepackage{verbatim}

\usepackage{amsthm}
\usepackage{color}

\usepackage{algorithm}
\usepackage{algorithmic}
\usepackage[algo2e]{algorithm2e} 
\usepackage{todonotes}

\usepackage{hyperref}

\usepackage[utf8]{inputenc}

\usepackage{fancyhdr}

\begin{document}

\mainmatter  

\title{Compressing Low Precision Deep Neural Networks Using Sparsity-Induced Regularization in Ternary Networks}

\titlerunning{Compressing Low Precision Deep Neural Networks }

%
\author{Julian Faraone*$^{\S}$%
\and Nicholas Fraser$^{\S}$* \and Giulio Gambardella$^{\S}$\and Michaela Blott$^{\S}$\and\\
Philip H.W. Leong*}
\authorrunning{Faraone et. al}

\institute{ *School of Electrical and Information Engineering, The University of Sydney, Australia\\ $^{\S}$Xilinx Research Labs, Dublin, Ireland\\
\mailsa\\
}

%
%

\toctitle{Lecture Notes in Computer Science}
\tocauthor{Authors' Instructions}
\maketitle

\vspace{-15pt}
\begin{abstract}
A low precision deep neural network training technique for producing sparse, ternary neural networks is presented. The technique incorporates hardware implementation costs during training to achieve significant model compression for inference. Training involves three stages: network training using L2 regularization and a quantization threshold regularizer, quantization pruning, and finally retraining. Resulting networks achieve improved accuracy, reduced memory footprint and reduced computational complexity compared with conventional methods, on MNIST and CIFAR10 datasets. Our networks are up to 98\% sparse and 5 $\&$ 11 times smaller than equivalent binary and ternary models, translating to significant resource and speed benefits for hardware implementations.
\end{abstract}

\section{Introduction}
Deep Neural Networks (DNNs) have revolutionized a wide range of research fields including computer vision ~\cite{CompVision} and natural language processing ~\cite{NLP}. However, along with excellent predicition capabilities, the state-of-the-art architectures are both computationally and memory intensive due to their vast number of model parameters. Ultra-low precision DNNs replace most floating point arithmetic with bitwise or addition operations which greatly reduces computational complexity and power consumption. These representations also significantly reduce hardware complexity and memory bandwidth, allowing implementations of state-of-the-art architectures on constrained hardware environments. As a result, there's been a growing interest in specialized hardware solutions for ultra-low precision DNNs and specifically, Binarized Neural Networks (BNNs) ~\cite{Binaryconnect}, ~\cite{Binarynet}, and Ternary Neural Networks (TNNs) ~\cite{TWN}. These networks constrain either weights alone or weights and activations, leading to extremely efficient hardware implementations. In the present work, we enhance the inherent sparsity of TNNs whilst maintaining the advantages of multiplierless computations. We use similar Convolutional Neural Networks (CNNs) to \cite{IntelSparse} for CIFAR10 classification and achieve similar accuracies, although their network has a full precision 1st layer compared to our ternary weights. Regularization techniques and reduced precision weight representations have been extensively studied for compression, acceleration and power minimization. Many efforts have concentrated on building efficient computational structures from floating point networks through sparse weight representations and quantization ~\cite{DeepCompression}, ~\cite{learning_prune}. Such networks still require fixed-point multiply-accumulate operations which limits power savings and speed. Instead of considering sparsity and reduced precision separately, we explore sparse TNNs which don't require multiplies in any layers. Pruning the fully connected layers of BNNs and TNNs was proposed in ~\cite{prune_low} to reduce the number of model parameters for efficient hardware implementations. We prune all layers and focus on inference acceleration. With recent breakthroughs in low precision deep learning, specialized hardware solutions have been increasingly investigated. FINN implements scalable BNN accelerators on FPGAs~\cite{FINN} and we use this framework to explore performance advantages of sparse TNNs. 

In this paper we propose a three-stage training approach for TNNs which is able to reduce hardware costs for inference. Firstly, the network is trained using L2 regularization and a quantization threshold regularizer, secondly we use quantization pruning whereby the sparsity pruning threshold is the same as the quantization threshold and thirdly we retrain the network. During training, the network learns in a sparse environment. This has significant benefits as we can determine the sensitivity of the weights to sparsity regularizers, i.e. measure the sensitivity of weights in different layers and advantageously utilize a quantization pruning method. 
The contributions of this paper are thus as follows:
\begin{itemize}
	\item The first reported low-precision training method which minimizes hardware costs as part of the objective function. This uses a quantization threshold regularizer and L2 regularization to encourage sparsity during training.
	\item A layer-based quantization pruning technique which utilizes sparsity information obtained during training.
	\item A quantitative comparison of our proposed sparse TNN with state-of-the-art multiplierless networks in terms of accuracy, memory footprint, computational requirements and hardware implementation costs.
	\item We achieve between 2 and 11x compression. For memory bound hardware architectures, this would directly translate into speed-up.
\end{itemize}
\vspace{-14pt}
\section{Sparse TNN Training}
The key idea in this work is to introduce sparsity in TNN weight representations through regularization. TNN training consists of real-valued weight parameters, $w_r$, which are quantized deterministically to $w_q$ using a quantization threshold, $\eta$. 
\begin{align}
w_q= 
\begin{cases}
\begin{aligned} 
& \text{$1$} \quad  \text{ if }\quad w_r > \eta\\
& \text{$0$} \quad  \text{ if }\quad   \mathbf{-\eta} \leq w_r \leq \eta\\
&  \text{$-1$} \quad  \text{ if }\quad  w_r < -\eta
\end{aligned}
\end{cases}
\end{align}
For the forward path, $w_q$ is computed and used for inference. For the backward path, the gradients are computed with $w_q$ and parameter updates are then applied to $w_r$. In training DNNs, generally many values for $w_r$ can achieve the same training loss and regularization techniques incorporate a preference for certain weight representations. We use several regularization techniques to minimize the number of nonzero parameters and induce sparsity. Our regularization scheme considers the hardware costs not only during the fine-tuning stage, but also during training.
\vspace{-10pt}
\subsubsection{Quantization Threshold Regularization}
Deterministic rounding requires partitioning the $w_r$ weight space by setting a threshold hyperparameter $\eta$. Typically different values for $\eta$ are set for different assumptions made on $w_r$. To uniformally partition the weight space $\eta = 0.33$ ~\cite{prune_low} or to minimize quantization error $\eta = 0.5$. In our case we increase $\eta$ to make 0's consume a large portion of the weight space (upto 95\%) which induces a similar sparsity effect to L1 regularization. However, L1 regularization has a continous shrinkage effect which induces sparsity amongst all  $w_r$ but not necessarily $w_q$. Increasing the threshold on the other hand, induces sparsity directly amongst $w_q$ and parameter updates for $w_r$ are either penalized or rewarded based purely on the gradients. An example of the regularization effect is shown in Figure 1(a) . By partitioning the weight space by 90\%, the network is initialized with high sparsity and takes longer to converge than training with a uniformally distributed weight space. 
\begin{figure}
	\vspace{-22pt}
\centering
\begin{subfigure}
	\centering
	\includegraphics[width=2.2in]{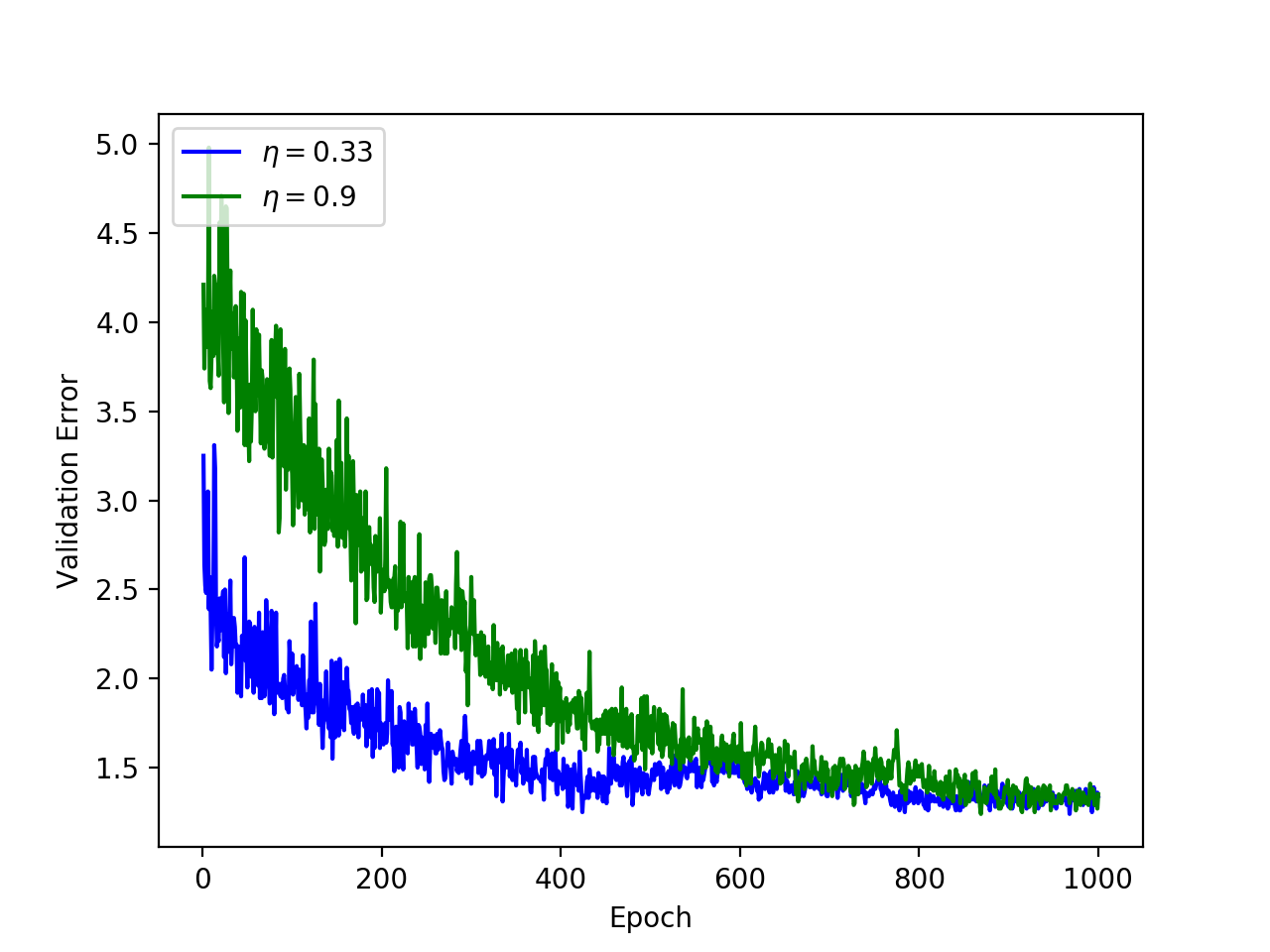}
	\label{fig:Conv}
\end{subfigure}%
\begin{subfigure}
	\centering
	\includegraphics[width=2.5in]{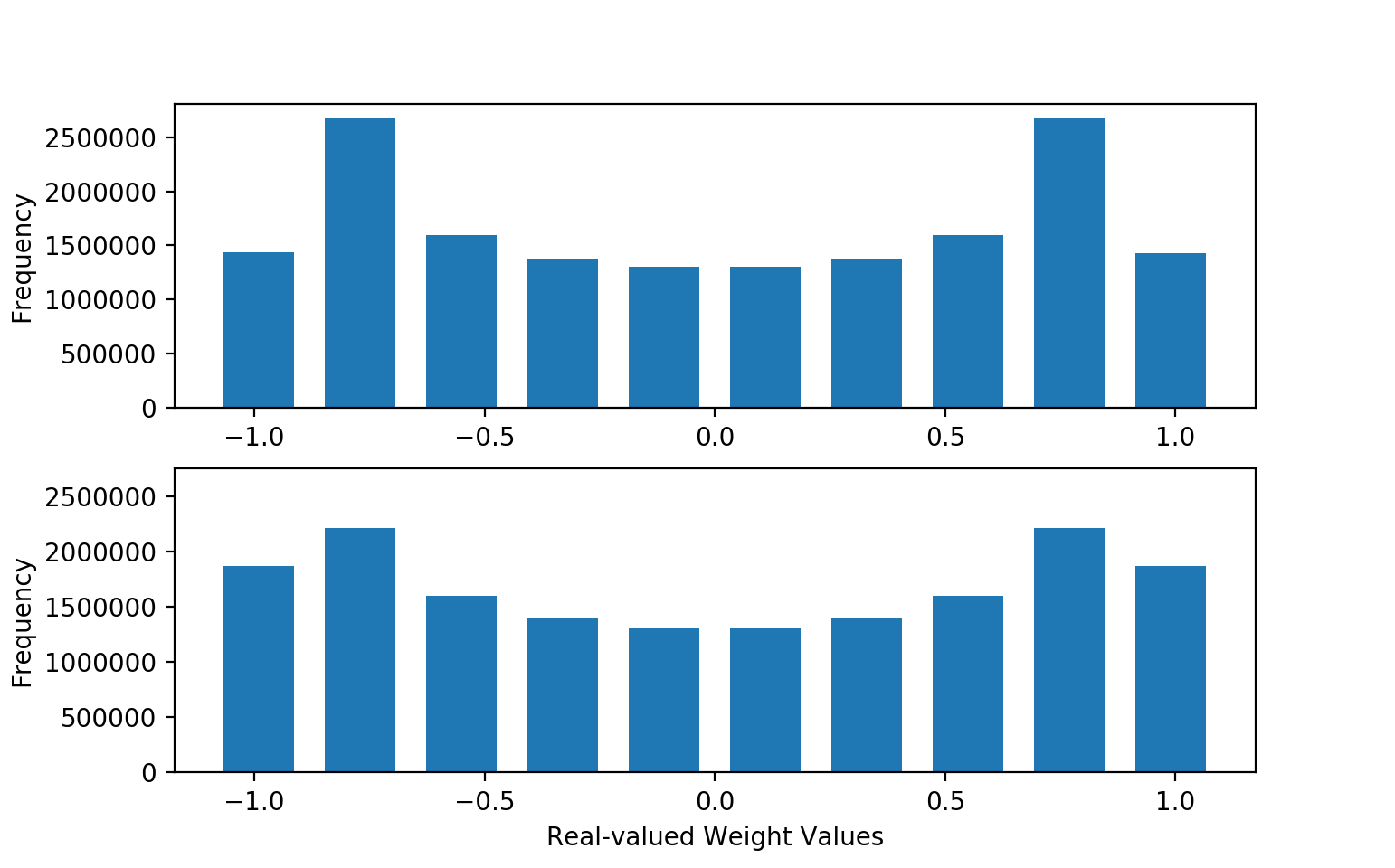}
	\vspace{-10pt}
	\caption{(a)Validation Error Convergence on MNIST. (b) Weight distribution for $w_r$ for MLP Layer in MNIST training. With L2 regularization (top) and without (bottom), for $\eta = 0.9$.}
	\label{fig:Weight_dist}	
\end{subfigure}
\vspace{-20pt}
\end{figure}
\vspace{-4pt}
\subsubsection{L2 Regularization}
In traditional TNN training the cost function $C$ can be represented as the average loss $L_i$ over all training examples $n$:
\begin{align}
&C(w_q) = \frac{1}{n}\sum_{i=1}^{n} L_i(w_q)
\end{align}
L2 regularization has the property of penalizing peaky weights to generate a more diffused set of weights. We add L2 regularization, as a function of the quantized weights, directly into the cost function to penalize nonzeros and induce sparsity:
\begin{align}
& C(w_q) =  \underbrace{\frac{1}{n}\sum_{i=1}^{n} L_i(w_q)}_{\text{data loss}}\quad+\underbrace{\lambda R(w_q)}_{\text{regularisation loss}}
\end{align}
where the regularization term is the quadtratic penalty over all parameters,
\begin{align}
& R(w_q) = \frac{1}{2}w_q^2
\end{align} 
and the gradient contribution from the regularization term becomes:
\begin{align}
& \frac{dC(w_q)}{dR(w_q)} = \lambda w_q
\end{align}
where $\lambda$ is the regularization strength hyperparameter. With L2 regularization, each epoch becomes a greedy search to reduce hardware costs as only the corresponding $w_r$ for each nonzero in $w_q$ is penalized by $\lambda$. From (5) and (1) it is evident the regularization term will only effect the corresponding parameter updates on $w_r$ for nonzero $w_q$. This is desirable when used in conjunction with a large $\eta$ as peaky weights (close -1 and 1 in this case) are more likely to be pulled below the threshold for a given regularizartion strength. Also, it avoids L2 regularization from continually penalizing weights, making them stuck at low values. This allows for weight values which are penalized in earlier training epochs, to then be more easily recovered through parameter updates if required later in training. As seen in Figure 1(b), under L2 regularization the frequency shrinks for weight values closer to -1 and 1. It is also evident that many weights clump around values closer to the threshold of 0.9.
\vspace{-10pt}
\subsubsection{Quantization Pruning}
L2 and threshold regularization achieve a certain sparsity before accuracy starts to degrade. This is addressed via quantization pruning by utilizing weight sensitivity information after the initial training phase and eliminating a subset of weights in $w_r$ which all quantize to zero. We then retrain the network, using a masking vector $w_m$ which sets the pruned weights to zero:
\begin{align}
{w_m} = 
\begin{cases}
\begin{aligned} 
&   {1}\quad  \text{ if }\quad  w_r < -\sigma  \\
&   {0}\quad  \text{ if }\quad   -\sigma \leq w_r \leq \sigma\\
&   {1}\quad  \text{ if }\quad  w_r > \sigma
\end{aligned}
\end{cases}
\end{align}
In sensitivity pruning, the sparsity hyperparameter $\sigma$ is optimized by setting different sparsities for different layers. Depending on the type and order of layer, they have a different sensitivity to pruning. In our method, by forcing sparsity through regularization during training, the gradient descent minimization process converges on the inherent sparisity sensitivity of each layer. We then utilize the ratio of zeros in each layer from the first training phase by pruning only $w_r$ below or equal to the quantization threshold. 
\begin{align}
& \sigma \leq \eta
\end{align}
Weight initialization for retraining then becomes the elementwise multiplication $w_{r_2}$.
\begin{align}
& w_{r_2} = w_{r_1} \odot w_m  
\end{align}
For retraining, $w_{r_2}$ is updated but the pruned weights are fixed at zero. Also, the threshold is set to the same value as in the initial training phase.
\vspace{-10pt}
\subsubsection{Weight Representations}
When implementing TNNs for inference on computer hardware, the real valued weights in $w_r$ are discarded and only $w_q$ is stored . In order to demonstrate the benefits of the sparse nature of these networks, we use two different compression techniques which can be utilized for different embedded device and specialized hardware applications depending on memory and resource requirements. Due to the high data regularity of the weight representations, storing all the ternary weights as 2-bits is not necessary. To conveniently store the unstructured sparse weight values, we use two compression methods. The first is Run Length Encoding (RLE), which stores only the index differences between each nonzero and also a sign bit which defines the type of operation. In our second method we use Huffman Coding (HC) on the index differences to assign variable length codewords whereby the most frequently occuring indexes are represented with shorter length codes and vice versa. HC has higher complexity for its decoder implementation and a higher compression rate than RLE.
\vspace{-10pt}
\subsubsection{Algortihm}
Algorithm~\ref{alg:one} is the compression process and consists of four parts. Part 1) respresents typical TNN training and additionally requires hyperparameters $\lambda$ and $\eta$ to be set as in Algorithm 2. In Part 2) the masking vector is computed and used for retraining in Part 3). After the network is trained, the real-valued weights are discarded and the quantized weights are encoded for Part 4). Outputs and inputs for each layer are represented by $y$ and $x$ respectively; $b$ is the bias term (if applicable); L is the learning rate; and CGU is compute gradient updates.

\vspace{-25pt}
\begin{minipage}[t]{5cm}
	\begin{algorithm}[H]
		\caption{}
		\label{alg:one}
		\begin{algorithmic}
			\STATE {\bfseries 1. Train} 
			\STATE Set $\lambda$ and $\eta$ for sparsity requirements and implement Algorithm ~\ref{alg:two}
			\STATE {\bfseries 2. Prune} 
			\STATE Compute $w_m$ \textbf{with} $\sigma = \eta$
			\STATE {\bfseries 3. Retrain}
			\STATE Keep $\lambda$ and $\eta$ the same
			\STATE Repeat Step 1. \textbf{with} $w_{r_{2}} = w_{r_{1}} \odot w_m$ and  $\lambda, \eta$
			\STATE {\bfseries 4. Encode } 
			\STATE Apply HC or RLE on resuting $w_q$ 
		\end{algorithmic}
	\end{algorithm}
\end{minipage}%
\begin{minipage}[t]{5cm}
	\vspace{0pt}
	\begin{algorithm}[H]
		\caption{}
		\label{alg:two}
		\begin{algorithmic}
				\STATE \textbf{-Forward Pass:}
			\FOR{each weight layer p}
			\STATE $w_{q_{p}} = Q(w_{r_{1_{p}}})$ \textbf{with threshold} $\eta$ \\
			\ENDFOR
			\FOR{each layer i in range(1,N) }
			\STATE Compute $y_{i}$ \textbf{with} $w_{q}$, $x_{i}$\\ 
			\ENDFOR
			\STATE \textbf{-Backward Pass:}
			\STATE Compute cost: $C(w_{q})$ \textbf{with} $y_{N}$, $\lambda$\\
			\FOR{each weight layer j }
			\STATE{ CGU: $g_{1} = \frac{dC(w_{q_{j}})}{dw_{q_{j}}} + \lambda w_{q_{j}}$ \\
				CGU:  $g_{2} = \frac{dC(w_{q_{j}})}{db_{j}}$\\
				Updates: $w_{r_{1_{j}}} = w_{r_{1_{j}}} - Lg_{1}$ \\
				$b_{j} = b_{j} - Lg_{2}$
			} 
			\ENDFOR
		\end{algorithmic}
	\end{algorithm}
\end{minipage}
\vspace{-15pt}
\section{Sparsity and Networks}
\vspace{-3pt}
We evaluate our training methods on two image classification benchmarks, MNIST and CIFAR10. We apply our training technique and compare directly against results from BinaryConnect ~\cite{Binaryconnect} and BinaryNet ~\cite{Binarynet}. BinaryConnect uses floating point ReLu activation functions and BinaryNet uses binary activation functions. Their results are represented in Figures 3 \& 5 as 'model-a-b' where a is the weight bitwidth and b is the activation bitwidth (bitwidth = 32 is for floating point, bitwidth = 1 is for binary and bitwidth = 2 is for ternary equivalents of these architectures with a uniformally distributed weight space). Our results are reported as TNN with resulting sizes represented as x/y which represents the sizes after encoding in RLE/HC respectively. For all results, we report the number of weight parameters (Params) in millions, percentage of zero-valued parameters, the error-rate and size of the network in megabytes (MB). In all our models we used only one pruning iteration except for the MLP with floating point activations for which we used two iterations. 
\vspace{-10pt}
\subsubsection{MNIST}
The MNIST dataset consist of 70k 28$\times$28 images of grey-scale handwritten digits. The networks used for classification consist of 3 hidden layers of 4096 neurons for the network with binary activations and 1024 neurons for the network with floating point activations. We train the network for 1000 epochs and choose the network which produces the best validation error rate. We first analyse the effect of quantization pruning on the MNIST dataset for different threshold settings. No L2 regularization is used in these numbers in order to focus on the effect of different pruning thresholds. Setting a higher threshold allows for more aggressive pruning at the threshold. The results are displayed in Figure 2 and although other threshold settings achieve similar accuracies, setting both $\eta=\sigma=0.9$  achieves significantly more sparsity. At this setting we can prune 80\% of weights without pruning away any nonzeros. It is evident that pruning nonzeros impinges on the network performance and hence the quantization threshold is an effective indicator for which weights can be pruned. Pruning at a lower sparsity threshold maintains accuracy benefits, although results in more nonzeros and to highly sparsify the network, it would have to be repeatedly pruned. This could require several iterations and take days/weeks as each training iteration takes days itself. The results are displayed in Figure 3. Using the network with binary activations produces up to 97.6\% sparsity and over 5$\times$ compression over its binarized network (BNet) with better accuracy and approximately 11$\times$ its ternary network with the same accuracy. The network with a floating point activation function, achieves 92.8\% sparsity and 3.5$\times$ compression over its binarized equivalent network. For these networks we used $\eta = 0.9$
\begin{figure}[t]
\centering
\setlength\tabcolsep{2pt}
\begin{minipage}{0.42\textwidth}
\centering
\begin{tabular}{lcccr}
		\hline
		\hline
		$\eta$ & $\sigma$ & Pruned & Error-rate & Nonzeros \\
		\hline
		\hline
		0.9 & 0.95 & 91\% & 1.08 & 1,220,468 \\
		$\mathbf{0.9}$   & $\mathbf{0.9}$&  $\mathbf{80\%}$&  $\mathbf{0.92}$& $\mathbf{1,863,521}$\\
		0.9 & 0.65 & 50\% & 0.96 & 3,826,912 \\
		\hline
		0.7   &0.8& 76\%& 0.98& 3,595,898 \\
		0.7  & 0.7 & 64\% & 0.91& 5,243, 764\\
		0.7 & 0.58& 50\% & 0.92 & 6,863,798 \\
		\hline
		0.5   &0.9& 89\%& 1.14& 2,448,073 \\
		0.5   &0.75& 74\%& 1.04& 5,416,539 \\
		0.5   &0.5& 74\%& 0.98& 10,396,476 \\\hline
\end{tabular}
	\caption{Quantization Pruning for TNN (Binary Activations) on MNIST, without L2 regularization}
	\label{fig:MNIST_sparsity}
\end{minipage}%
\hfill
\begin{minipage}{0.5\textwidth}
\centering
			\begin{tabular}{lcccr}
				\hline
				\hline
				Model & Params & Zeros & Error-rate & Size (MB) \\
				\hline
				\hline
				MLP-2-1   &36.4& 54\%& 0.92& 9.12 \\
				MLP-1-1    & 36.4&  0\%&  0.96& 4.56\\
				\hline
				TNN    & 36.4& 97.6\%& 0.93 & 0.83/1.59 \\
				\hline
				\hline
				MLP-2-32 & 2.91& 34\%& 1.23 &  0.72\\
				MLP-1-32 & 2.91 & 0\%& 1.29 & 0.36 \\
				\hline
				TNN   & 2.91& 92.8\%& 1.22 & 0.07/0.11\\
				\hline
			\end{tabular}
\caption{Classification accuracies for Sparse TNNs for MLPs on MNIST with L2 regularization and pruning}
\label{fig:mnist_res}
\end{minipage}
\vspace{-10pt}
\end{figure}
\vspace{-10pt}
\subsubsection{CIFAR10}
The CIFAR10 dataset is benchmark dataset consisting of 32$\times$32 colour images with 10 categories. We use a VGG-derivative architecture inspired by BinaryConnect ~\cite{Binaryconnect}. From Figure 5, we see that there is an improvement in accuracy and/or compression for both networks in contrast to their binarized and ternary equivalents. The convolutional layers are less robust to the threshold regularizer and hence we set a lower value for the convolutional layers $\eta_1 = 0.8$ and a higher value for the fully connected layer $\eta_2 = 0.9$. We show the accuracy and sparsity relationship in Figure 4(a) for varying thresholds and show that threshold regularization improves accuracy. The leftmost point is the fully dense binarized network where $\eta=0$ and as we introduce the threshold regularization, the error-rate drops by up to 1.4\% as the sparsity is increased. In Figure 4(b), we plot the percentage of nonzeros for each layer in the CNN for varying values of the threshold regularizer. By increasing the threshold, the robustness of each layer under sparsity becomes more prominent. For most of the networks, the first two convolutional layers are the most sensitive to sparsity and consist of around 80\% nonzeros and the last convolutional and first fully connected layers are the least sensitive. These are similar conclusions to \cite{learning_prune} who pruned each layer independantly to determine their sparsity sensitivity. In our case, the network learns these sensitivities by training in sparse environments. This is advantageous as efficient sparsity parameters are determined for any layer type or order and don't require a hyperparameter search. Varying the threshold provides sensitivity information for the sparsity of each layer and quantization pruning takes advantage of this by pruning each layer according to the threshold and hence these ratios. 
\begin{figure}[t!]
	\vspace{-15pt}
	\includegraphics[width=2.4in]{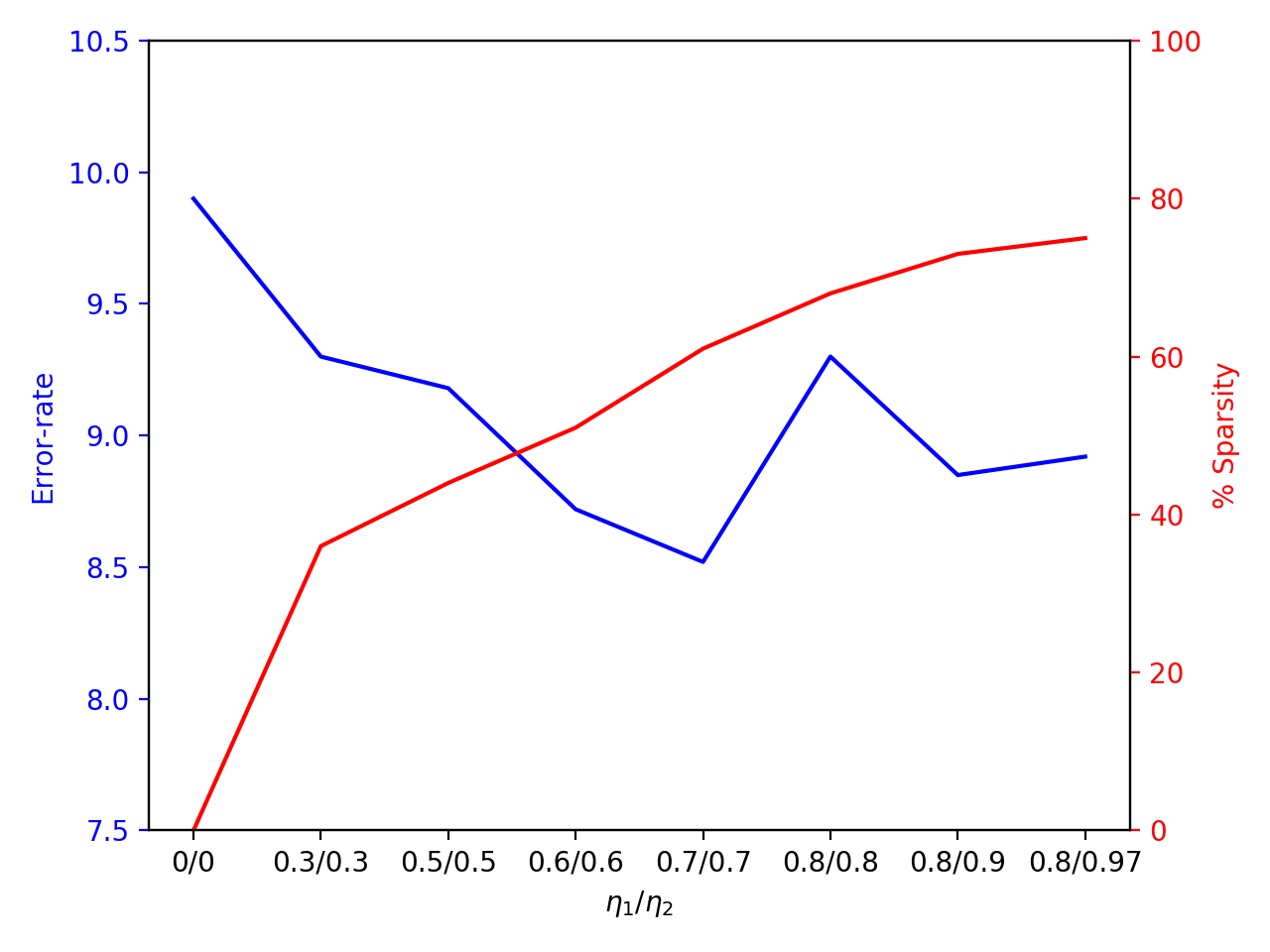}
	\label{fig:cifar10_accuracy}
	\includegraphics[width=2.6in]{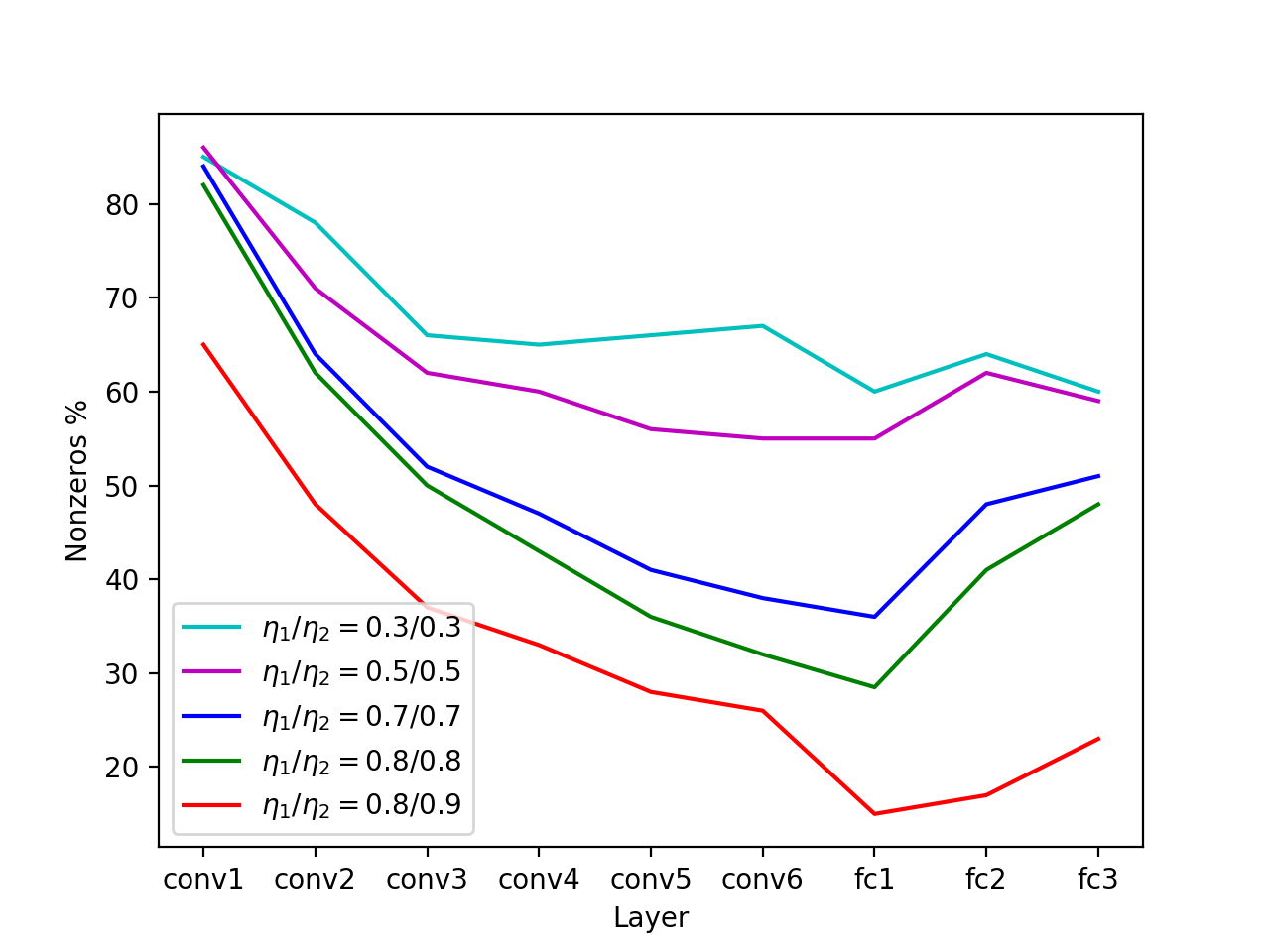}
	\vspace{-15pt}
	\caption{(a) Accuracy vs Sparsity for CIFAR10 with varying $\eta$, without pruning. (b) Per Layer Sparsity for different threshold values on CIFAR10, no regularization or pruning}
	\label{fig:per_layer_sparsity}
	\vspace{-1pt}
\end{figure}
\begin{figure}[t]
	\centering
	\setlength\tabcolsep{2pt}
\begin{minipage}{\textwidth}
	\begin{minipage}{0.5\textwidth}
		\centering
					\begin{tabular}{lcccr}
						\hline
						\hline
						Model & Params & Zeros & Error-rate & Size (MB) \\
						\hline
						\hline
						VGG-2-1   &14.02& 65\%& 11.2& 3.52 \\
						VGG-1-1    & 14.02&  0\%&  11.4& 1.76\\
						\hline
						TNN   & 14.02& 92.3\%& 10.8 & 0.88/1.05 \\
						\hline
						\hline
						VGG-2-32 & 14.02& 35\%& 9.2 &  3.52\\
						VGG-1-32 & 14.02 & 0\%& 9.9 & 1.76 \\
						\hline
						TNN   & 14.02& 90.1\%& 9.6 & 0.96/1.22 \\
					\end{tabular}
				\vskip 0.05in
				\vspace{-10pt}
		\caption{Classification accuracies for Sparse TNNs for CNNs on CIFAR10 with L2 regularization and pruning}
		\label{fig:cifar10_res}
\end{minipage}%
\hfill
\begin{minipage}{0.46\textwidth}
\centering
	\includegraphics[width=2.4in]{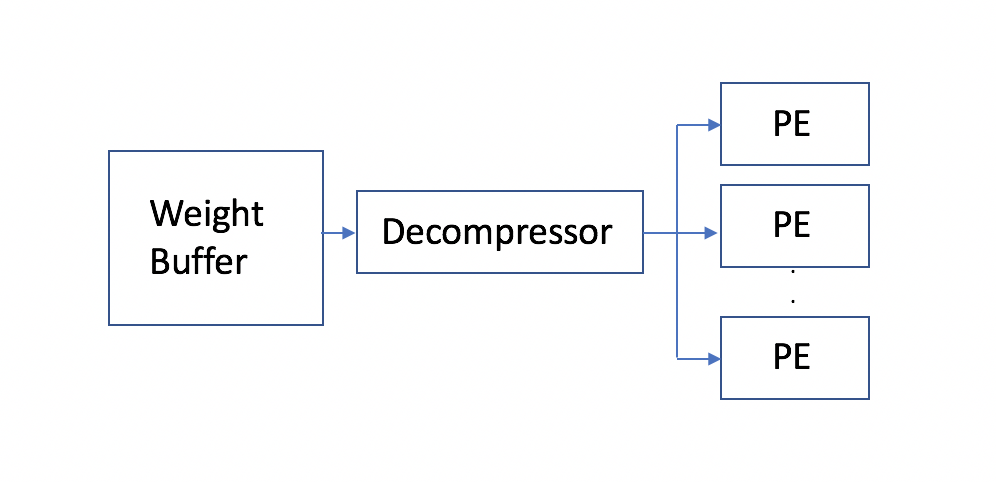}
	\vspace{-20pt}
	\caption{Diagram of Decompressor Feeding Multiple Processing Elements with Data Reuse}
	\label{fig:pe}
\end{minipage}
\end{minipage}
\vspace{-10pt}
\end{figure}
\section{Hardware Implications of Sparse TNNs}
\vspace{-5pt}
In this section, we explore the hardware implications of implementing TNNs with unstructured sparse data representations. Storing the weights in a compressed format, requires a decompressor which incurs some overheads on hardware designs. A fully parallel architecture would require a decompressor for every weight in the convolution or fully connected layer and decompressors in this case would consume significant amounts of resources. For sparse TNNs, we can take advantage of data reuse patterns which are present within convolution layers and fully connected layers (when batching is applied) to increase the ratio of processing elements (PEs) to decoders. When the sparsity of these networks is taken into consideration, the number of \textit{effective operations} (discussed later in this Section) increases the potential performance of TNNs to values well beyond those of BNNs. For conventional computing platforms (e.g., CPUs and GPUs), the main benefit of sparsity and compression is the increase in operational intensity that is achieved for a particular layer. Sequential processors, (such as CPUs) will also be able to benefit from the reduction in required operations per layer, as a result of the high sparsity of TNNs.
For parallel processors, (such as GPUs, FPGAs and ASICs) it is a lot more difficult to take advantage of this benefit due to the irregular data access patterns. We describe a hardware decompressor, a corresponding parallel architecture suitable for FPGAs and the potential performance of that architecture in terms of effective operations per second.
\vspace{-15pt}
\subsubsection{Hardware Decompressor}
Our proposed hardware decompressor iterates through a list of weights, stored in a sign--magnitude form in on-chip memory. In each cycle, the hardware decompressor outputs the complement of the sign bit to represent the weight value and adds the magnitude value to an internal counter, which is used to generate the address of the value to be accessed from the input vector. The RLE decoder consists of a counter which controls the address of the input to feed into the PE for computation. The resource and performance estimates given by Vivado HLS of the resultant hardware description are that the design can produce an address and a weight every cycle at 250 MHz while using 112 LUT resources on the FPGA.
\vspace{-15pt}
\subsubsection{A Sparse TNN Accelerator}
Three types of low precision networks are described in this paper:
1) networks with binary activations (VGG/MLP-1/2-1); and
2) networks with floating point activations (VGG/MLP-1/2-32).
For all networks, the predominant calculations for inference are multiply accumulate operations (MACs).
For type 1) networks, this corresponds to XNOR-popcount operations~\cite{Binarynet}, where a popcount is the number of set bits in a word.
For type 2) networks, this corresponds to an XNOR operation on the sign bit of a floating point value, followed by a floating point accumulate.
\vspace{-25pt}
\subsubsection{Accelerator Architecture}
Our proposed accelerator architecture is based on that generated by \textsc{FINN}~\cite{FINN}. In particular, we propose a design which has processing engines with a similar datapath to \textsc{FINN}.
To compute the input-weight matrix in specialized hardware implementations, typically a series or array of PEs are used to receive input data and a weight value to perform the multiply accumulate operations, as required for the datatype.
For the compressive format described in Section 2, these implementations require a decompressor between the weight matrix and the PE as represented in Figure 6. For type 1), we estimate resource usage on the roofline given by ~\cite{FINN}, which is reported to have an average cost of 5 LUTs for both an XNOR and popcount operation.%
\footnote{FINN quotes 2.5 LUTs per operation, which is multiplied by 2 to get LUTs / per MAC.}
For type 2), we estimate the resource usage by instantiating a Xilinx Floating Point 7.1 IP core addition module.
The peak throughput numbers are what can be achieved if 70\% of the LUTs or 100\% of the DSPs are used on the target device, a Xilinx KU115 running at 250 MHz. These are 46.4 TOPs for type 1) and 1.3 TOPs for type 2). The total KU115 resources are 663k LUTs and 5,520 DSPs.
\vspace{-2pt}
\subsubsection{Exploiting Sparsity Through Data Reuse}
\label{sec:data-reuse}
Convolutional layers require many operations on different input pixels to the same weight value.
Hence, we can utilize data re-use optimizations ~\cite{ScalingFINN} to instantiate a decompressor for a specific weight and calculate several MAC operations on different input pixels.
This greatly reduces the average resource usage of the decompressor per operation.
Similar optimizations can be utilized for the fully connected layers, whereby batching can be applied to allow a single weight to calculate several MAC operations across multiple input vectors.

Let us introduce a data re-use factor, $R$, which denotes the total amount of data re-use available in a particular layer.
For fully connected layers, $R=B$, where $B$ is the batch size.
For convolutional layers, $R=B\times P$, where $P$ is the number of output pixels in the output image.
Furthermore, our RLE decoder allows us to easily avoid calculating any zero valued weights.
In comparison to the benchmark BNNs, which have strictly dense weights, only the non-zero weight computations need to be calculated.
Our sparsity factor, then becomes a multiplier which significantly reduces our cost per operations and hence the regularization techniques discussed in the paper directly minimize hardware costs during training. To this end, we introduce an \textit{effective operation} cost, given by: $C_e = \gamma*(C_{op} + C_{d}/R)$,
where $\
ma$ is ratio of non-zero weight values to total weights in the layer,
$C_{op}$ is the proportion of the KU115 which is utilised by a single operation
and $C_{d}$ is the proportion of the KU115 which is utilised by the decoder.%
\footnote{Assuming 70\% of the LUTs and 100\% of the DSPs can be utilised for compute.}
An \textit{effective throughput} can then be calculated as: $T_e = 1/C_e * 250 \mathrm{MHz}$.
Figures 7(a) \& 7(b) show the effective throughput of type 1) \& 2) networks respectively,
while varying $\gamma$ and $R$. The horizontal lines represent the benchmark BNN  networks, MLP and VGG (VGG is labelled as CNN in figures) from the reults in Figures 3 \& 5  and the other percentages reperesent networks of the same type with varying sparsities.
Note that these are theoretical peak values and further overheads are likely for all datapoints when they are implemented in a real system. For type 2), a lower sparsity factor is required to improve on the benchmark throughput as these operations are more expensive and hence every zero weight has a greater hardware benefit than for the type 1). 
\vspace{-10pt}
\begin{figure}[t!]
	\vspace{-5pt}
	\includegraphics[width=2.3in]{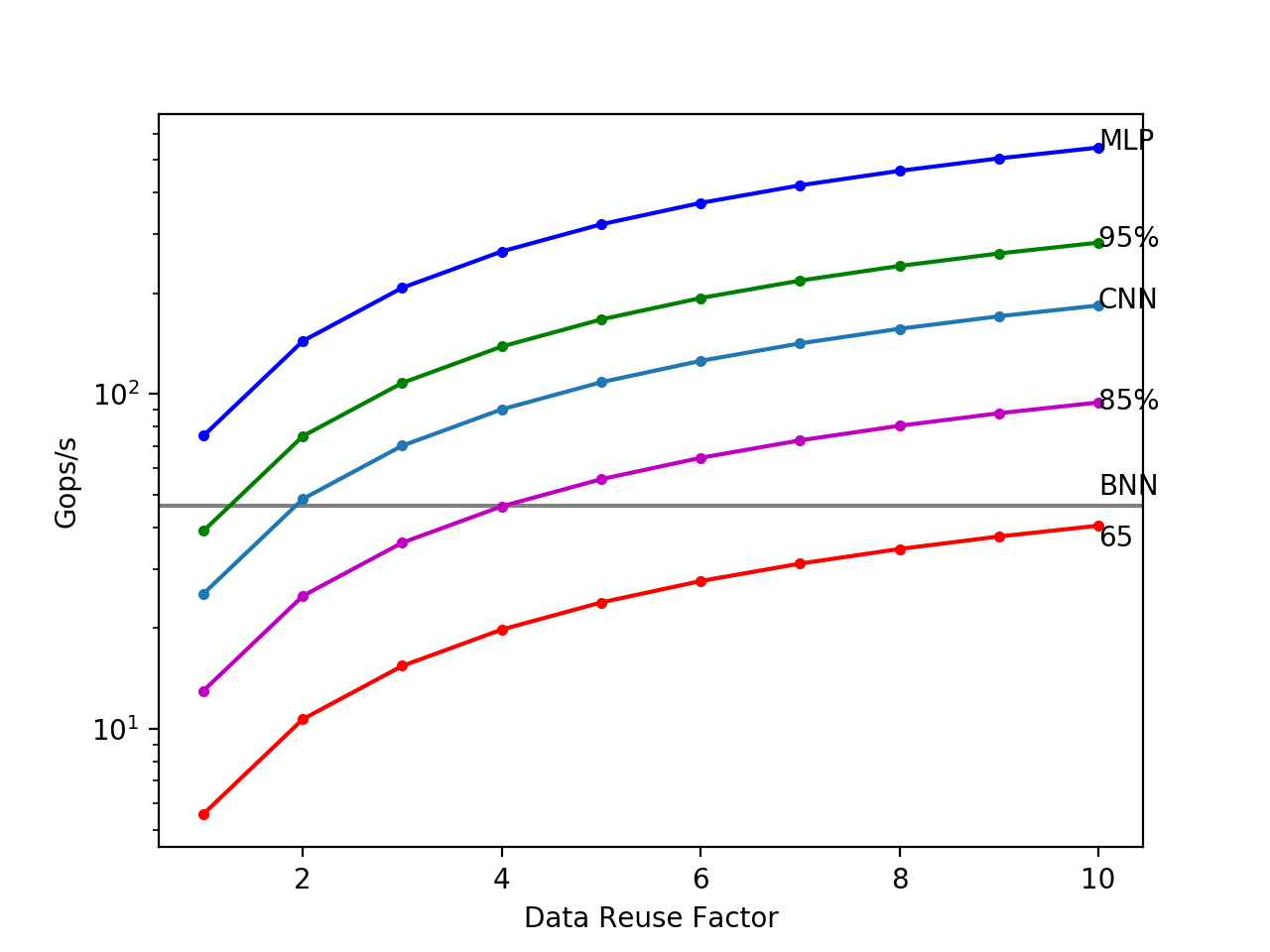}
	\label{fig:TNN-effective-throughput}
	\includegraphics[width=2.3in]{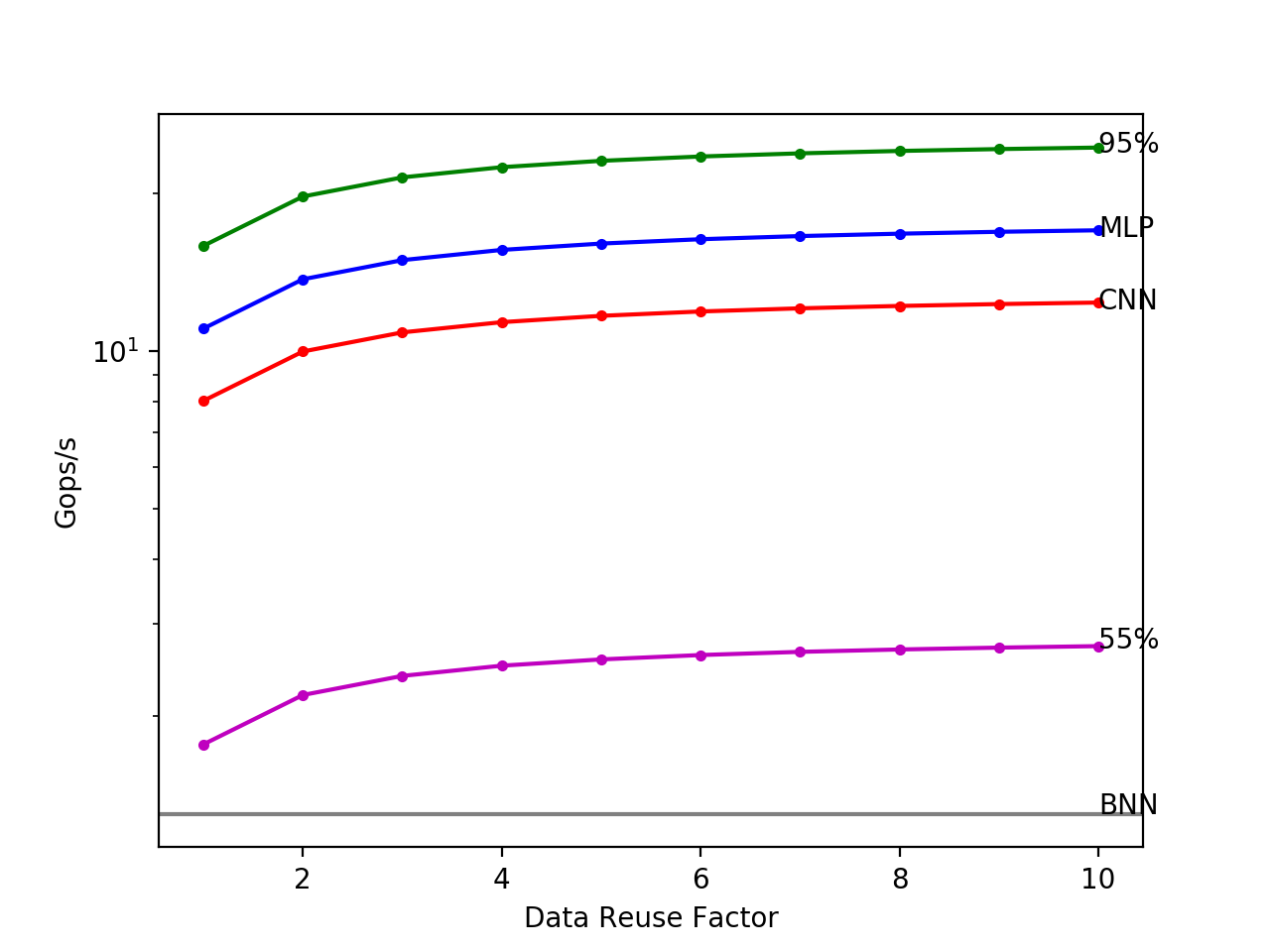}
	\label{fig:TCON-effective-throughput}
	\vspace{-10pt}
	\caption{(a) Effective throughput: BNNs vs TNNs (type 3) while varying $\gamma$ and $R$. (b) Effective throughput: BNNs vs TNNs (type 1) while varying $\gamma$ and $R$. Note: VGG is labelled as CNN.}
	\vspace{-10pt}
\end{figure}
\section{Conclusion and Future Work}
This paper contributes to the applicability of Deep Neural Networks on embedded devices and specialized hardware. We introduce a TNN training method which uses a quantization threshold hyperparameter, complemented by L2 regularization and quantization pruning to substantially reduce the memory requirements and computational complexity. This was shown using different network topologies on the MNIST and CIFAR10 benchmarks. Future work in ths area will look into extending this quantization technique to other low precision networks for more difficult datasets, improving accuracy whilst maintaining sparsity and also sparse TNN hardware accelerator designs.
\vspace{-20pt}
\subsubsection{Acknowledgements}
This research was partly supported under the Australian Research Councils Linkage Projects funding scheme (project number LP130101034) and Zomojo Pty Ltd.

\end{document}